\pdfoutput=1

\documentclass[11pt]{article}

\usepackage{acl}

\usepackage{times}
\usepackage{latexsym}
\usepackage{booktabs}
\usepackage{enumitem}
\usepackage{amsmath}
\usepackage{amssymb}
\setlist{nolistsep,noitemsep,label=$\triangleright$}

\usepackage[T1]{fontenc}

\usepackage[utf8]{inputenc}

\usepackage{microtype}

\usepackage{graphicx}
\usepackage{svg}

\newcommand{\dataset}{ASL STEM Wiki} 
\newcommand{\datasetshort}{ASL STEM Wiki} 
\newcommand{\challenge}{automatic sign suggestion} 

\newcommand{\ky}[1]{\textcolor{violet!70!white}{[Kayo: #1]}}
\newcommand{\todo}[1]{\textcolor{red}{[TODO #1]}}
\newcommand{\hal}[1]{\textcolor{magenta}{[Hal: #1]}}
\newcommand{\db}[1]{\textcolor{olive}{[Danielle:#1]}}
\renewcommand{\ky}[1]{}
\renewcommand{\todo}[1]{}
\renewcommand{\hal}[1]{}
\renewcommand{\db}[1]{}

\title{ASL STEM Wiki: Dataset and 
Benchmark for \\ Interpreting STEM Articles}

\author{Kayo Yin$^{\alpha}$
        \qquad  Chinmay Singh$^{\beta}$
        \qquad Fyodor O Minakov$^{\beta}$
        \qquad Vanessa Milan$^{\beta}$\\ 
        \textbf{Hal Daumé III}$^{\beta,\gamma}$
        \qquad \textbf{Cyril Zhang}$^{\beta}$
        \qquad \textbf{Alex X. Lu}$^{\beta}$
        \qquad \textbf{Danielle Bragg}$^{\beta}$\\\\
        $^{\alpha}$University of California, Berkeley
        \qquad $^{\beta}$Microsoft Research
        \qquad $^{\gamma}$University of Maryland\\
        \texttt{kayoyin@berkeley.edu}, \texttt{hal3@umd.edu}\\
        \texttt{\{chsingh,cyrilzhang,lualex,danielle.bragg\}@microsoft.com}
}

\begin{document}
\maketitle


\begin{abstract}
Deaf and hard-of-hearing (DHH) students face significant barriers in accessing science, technology, engineering, and mathematics (STEM) education, notably due to the scarcity of STEM resources in signed languages. To help address this, we introduce \dataset{}: a parallel corpus of 254 Wikipedia articles on STEM topics in English, interpreted into over 300 hours of American Sign Language (ASL). \dataset{} is the first continuous signing dataset focused on STEM, facilitating the development of AI resources for STEM education in ASL.
We identify several use cases of \dataset{} with human-centered applications. For example, because this dataset highlights the frequent use of \textit{fingerspelling} for technical concepts, which inhibits DHH students' ability to learn,
we develop models to identify fingerspelled words---which can later be used to query for appropriate ASL signs to suggest to interpreters.\footnote{Dataset and code: \url{https://www.microsoft.com/en-us/research/project/asl-stem-wiki}}

\end{abstract}

\section{Introduction}\label{sec:intro}

\begin{figure}
    \centering
    \includegraphics[width=\linewidth]{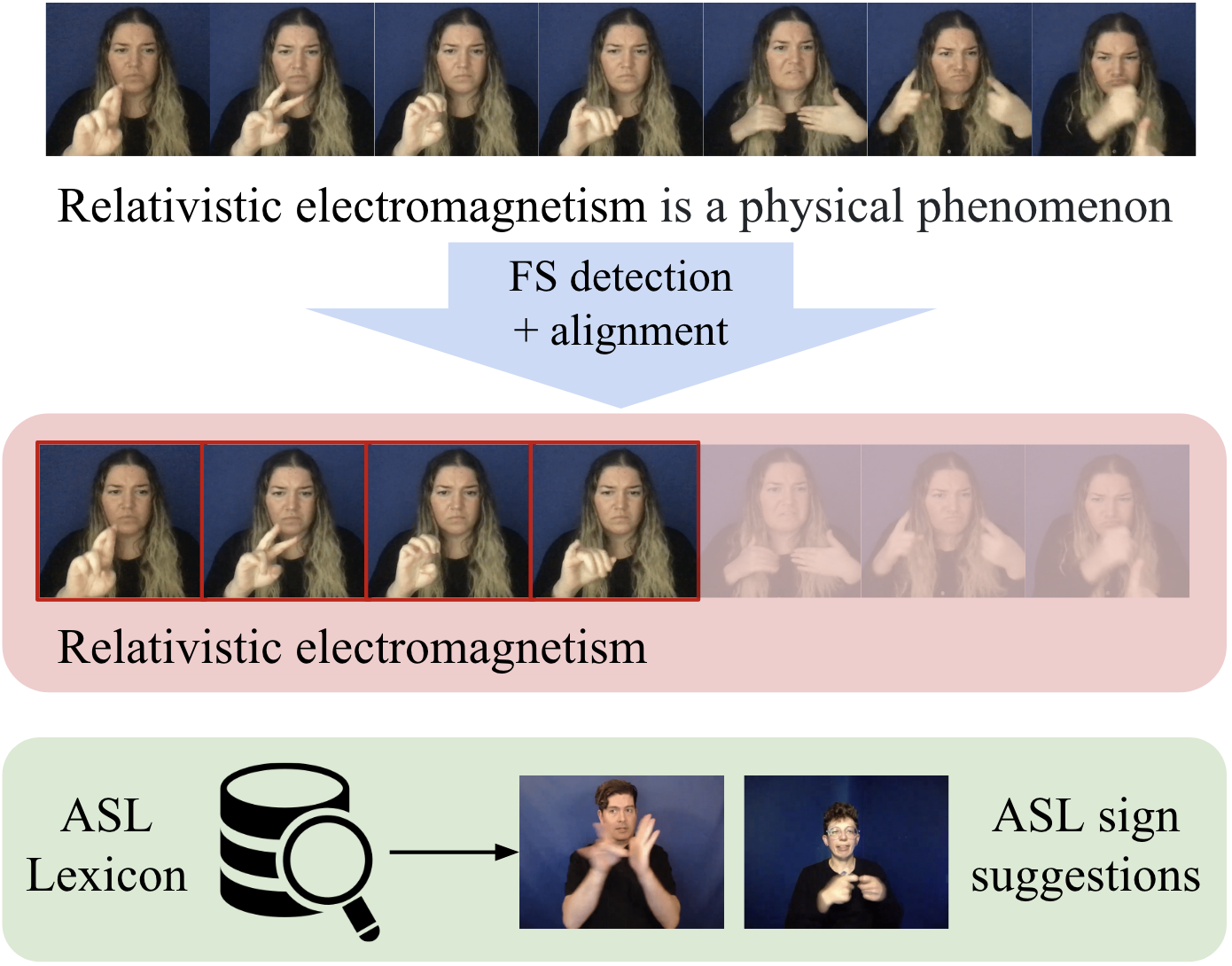}
    \caption{One use case of \dataset{} is automatic sign suggestion. Given an English sentence and a video of its ASL interpretation, the model detects all clips of ASL that contains fingerspelling (FS). Then, given the detected FS clip and the English sentence, the model identifies which English phrase in the sentence is fingerspelled in the clip. The English phrase can be used to query an ASL lexicon and suggest ASL signs.}
    \label{fig:task}
\end{figure}

American Sign Language (ASL) is the primary and the most accessible language for many deaf children in the U.S.
Despite the crucial importance of accessible education in ASL, deaf and hard of hearing (DHH) students often face significant barriers in accessing science, technology, engineering, and mathematics (STEM) education \cite{pagliaro2013math, traxler2000stanford}. Few general educational resources exist in ASL, and even fewer ASL resources exist for STEM content. 
The scarcity of STEM resources in ASL compounds challenges experienced by DHH students with limited English literacy \cite{lang2003web}.

\begin{figure*}
    \centering
    \includegraphics[width=\linewidth]{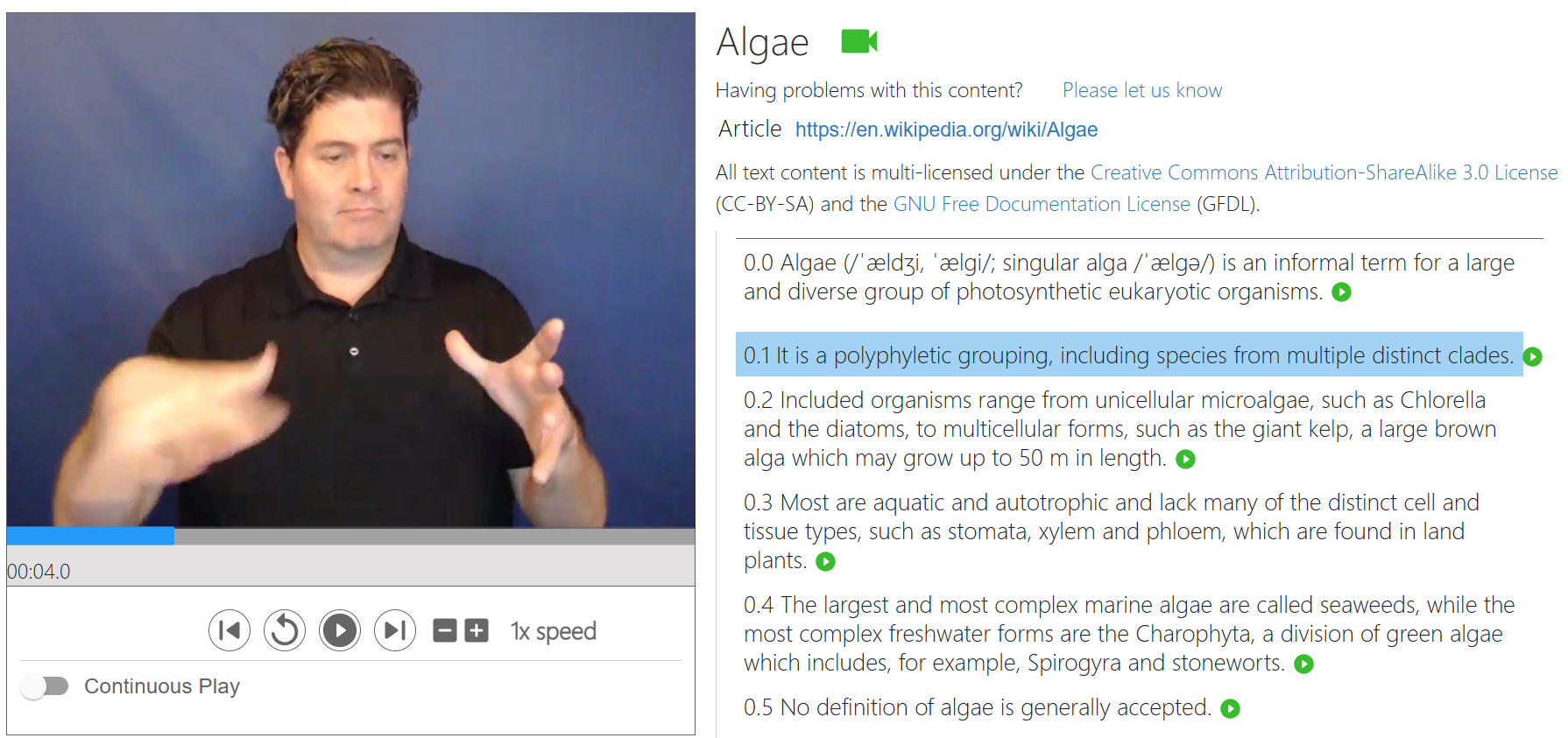}
    \caption{The bilingual resource used to both collect and display \dataset{}. The design was proposed in \citet{glasser2022asl}.
    Contributors select a sentence that they would like to interpret, which activates their webcam for recording. Consumers can read articles in English and access ASL interpretations for desired sentences.}
    \label{fig:resourceScreenshot}
\end{figure*}

In particular, the lack of standardization of STEM terminology in ASL creates obstacles for deaf education \cite{lang2007study}. 
However, despite the existence of ASL signs for STEM concepts proposed by deaf STEM educators or scientists, many of these signs have not yet been widely disseminated or adopted as standard vocabulary within the community. In the absence of conceptually accurate signs, interpreters often resort to suboptimal strategies \cite{lualdi2023advancing}. These include translating English words without considering the conceptual meaning (e.g.,  using the sign for ``intention'' to translate ``mathematical \textbf{mean}''); using a placeholder shorthand (e.g. the handshape letter ``M''), which carries no meaning on its own and must be remembered for each context; or \textit{fingerspelling} the term using a sequence of one-handed signs that each represents an English letter, instead of signing it (e.g., letter signs for M-E-A-N), which takes more time and forces the student to switch between ASL and English. These strategies often reference English and may hinder students from learning new concepts \cite{higgins2016development, enderle2020communicating}. 

To help develop better ASL STEM resources and AI models, we introduce the \textbf{\dataset{}} dataset (\autoref{fig:resourceScreenshot}): a parallel corpus of Wikipedia articles on STEM topics in English and American Sign Language (ASL). \dataset{} consists of 254 English Wikipedia articles on 5 STEM-related topics, interpreted into ASL by 37 certified interpreters, resulting in 64,266 sentences and over 300 hours of ASL video. 
It is the first continuous sign language dataset focused on STEM content, which introduces new AI modeling challenges. 
In \autoref{sec:dataset}, we explain how \dataset{} was collected and present exploratory dataset statistics. We also propose several possible use cases for this dataset informed by its unique characteristics. 


We present novel baseline models focused on addressing the high rate of fingerspelling in STEM content. Specifically, we tackle \challenge{} (\autoref{fig:task}; \autoref{sec:task}): given an English sentence and a video of its ASL interpretation, a model detects when the interpreter uses fingerspelling and suggests appropriate ASL signs to use instead. The motivations behind this challenge are two-fold: first, as described above, this challenge encourages development of tools that can aid human interpreters in producing higher-quality ASL interpretations of technical content, which in turn increases access to educational material in ASL and promotes the usage of technical ASL signs. Second, it creates a unique opportunity to drive progress and evaluate AI systems on ASL understanding. To train our baseline models, we use contrastive learning to leverage the large set of unlabeled ASL videos, and find that contrastive learning can mitigate annotation scarcity and improve fingerspelling detection IOU score by 47\% (\autoref{sec:model}).

Our primary contributions are:
\begin{enumerate}
    \item We present the first dataset of continuous STEM content in ASL (or any other signed language) and the largest dataset with signer consent. The dataset was recorded by certified interpreters, covers English STEM Wikipedia articles, and includes sentence alignment. 
    \item We provide fingerspelling detection and alignment baselines as first steps towards \challenge. We also outline several other new modeling challenges introduced by our continuous STEM dataset. 
    \item We provide the first evidence that self-supervised contrastive pretraining can improve fingerspelling detection. \end{enumerate}

\section{Background \& related work}\label{sec:bg}

\paragraph{STEM communication and education in American Sign Language.}
ASL is a relatively young language and is primarily used by deaf individuals who have historically been underrepresented in science. This in turn limits the growth of ASL in STEM fields \cite{cavender2010asl, lualdi2023advancing}. As a result, many STEM concepts do not have an agreed upon ASL sign \cite{lang2007study}. Deaf students often encounter alternative signs for the same term from different teachers or professionals, and signs generated on the fly are often phonologically or morphologically incorrect,
which further hinders ease of learning and scientific communication. Studies also show that deaf students report sign clarity as the top priority of teacher characteristics \cite{lang1993characteristics}.

To address challenges in STEM communication in ASL, several efforts have proposed mechanisms for signers to share and discuss ASL signs online \cite{cavender2010asl, reis2015asl} in an attempt to disseminate and standardize technical ASL signs. However, this does not fully address challenges for comprehension in interpreted ASL, especially in the educational setting. \citet{kurz2015deaf} find that deaf students score higher on scientific knowledge with direct instruction in ASL than with interpreted instruction. 

Interpreted ASL is also more prone to influence from English, such as using an English word order that would be unnatural in ASL; using the ASL sign for an English homonym instead of the sign that semantically matches the source word (e.g. ``protein'' in nutrition vs. biology), or defaulting to fingerspelling when the interpreter does not know the corresponding ASL sign. 
Since fingerspelling invokes an entirely separate language, the overreliance on fingerspelling is unlikely to help deaf students acquire a conceptual understanding of the term \cite{enderle2020communicating}, and deaf students report to prefer scientific concepts to be signed, rather than fingerspelled alone \cite{higgins2016development}.

\begin{table*}[t]
    \centering
      \resizebox{0.8\linewidth}{!}{
    \begin{tabular}{llllr}
    \toprule
     \textbf{Dataset}   & \textbf{Source \& Topic}  & \textbf{Signers} & \textbf{Consent?} & \# \textbf{Hours}  \\
     \toprule
     How2Sign \cite{how2sign}   & ``How-to'' YouTube &   Interpreter& Yes & 80  \\
      OpenASL \cite{openasl}   & Deaf YouTube &  Deaf \& interpreter & No & 288 \\
      YouTube-ASL \cite{youtubeasl}   & YouTube &   Unknown & No & 984  \\
     \midrule
    \dataset{} (Ours)  & STEM Wikipedia &  Interpreter & Yes & 316   \\
     \bottomrule
    \end{tabular}}
    \caption{Summary statistics of English-ASL continuous datasets. We report the source and topic of dataset content, the type of signers included, whether signers have consented to appearing in the dataset, and the number of hours of ASL video in the dataset.}
    \label{tab:datasets}
\end{table*}

\paragraph{ASL datasets.}
Several large datasets with continuous ASL have been collected recently to further the development of AI systems for ASL. We summarize existing ASL datasets in \autoref{tab:datasets}. How2Sign \cite{how2sign} consists of a parallel corpus of speech and transcriptions of ``How-to" instructional videos, and corresponding ASL interpretations, totaling over 80 hours of ASL videos. OpenASL \cite{openasl} gathers ASL videos and English captions from three Deaf YouTube channels 
and contains 288 hours of ASL videos. YouTube-ASL \cite{youtubeasl} consists of ASL videos and accompanying English captions extracted from YouTube 
and contains around 1000 hours of ASL videos. 
In contrast to the former two datasets that is composed fully of fluent or professional signing, YouTube-ASL may contain novice levels of ASL. 

Our dataset is the first dataset focused on continuous ASL in the STEM domain, which presents a valuable resource for deaf students and unique opportunities and challenges for modeling (\autoref{sec:dataset}). With over 300 hours of ASL content, it is also the largest ASL dataset where individuals appearing in the dataset have consented to use of their data, and where each sample contains professional ASL signing (since our dataset consists fully of certified ASL interpreters).


\section{\dataset{} Dataset}
\label{sec:dataset}


We now describe how we collected \dataset{}.  Our dataset is published under a license that permits use for research purposes. We also include the datasheet of \dataset{} in \autoref{sec:datasheet} with further details. \ky{Should we include this in camera-ready?}


\subsection{Text curation}

 We first selected a set of STEM-related Wikipedia articles to interpret. From a public Wikipedia download\footnote{\url{https://en.wikipedia.org/wiki/Wikipedia:Database_download}} accessed July 2020, 
 we filtered to popular\footnote{\url{https://en.wikipedia.org/wiki/Wikipedia:Popular_pages}} or important\footnote{\url{https://en.wikipedia.org/wiki/Wikipedia:Vital_articles}} articles, 
and selected STEM-related articles using topic modeling and manual validation.  
The final set contains 254 articles, including 113 articles in Science, 50 in Geography, 47 in Technology, 26 in Mathematics, and 18 in Medicine. 


We also selected a subset of five ``control'' articles, chosen to cover diverse topics and to be short. 
We use these control articles to verify interpreter quality, as all interpreters were asked to record these articles. The articles selected as control are: Acid catalysis, EDGE species, Hal Anger (person), Relativistic electromagnetism, and Standard score.

\subsection{Video collection}

With IRB and ethics review approval, we recruited 37 professional ASL interpreters to provide interpretations of the STEM-related Wikipedia articles described above. 
Videos were collected through a website, and on their first visit, contributors engaged in a consent process and were asked about basic demographics.

Of the 37 interpreters we recruited, $59\%$ report their gender as female, $30\%$ as male, $3\%$ other (and $8\%$ undisclosed), with ages in the 28--59 range ($\mu= 42$, $\sigma=8.6$). Eight identified as d/Deaf, 27 as hearing, one undisclosed, and one other. All interpreters are certified by the Registry of Interpreters for the Deaf or the Board for Evaluation of Interpreters.
All participants were compensated at a standard hourly remote interpreting rate. 

Participants provided interpretations using our custom implementation of the interface described in \citet{glasser2022asl} (\autoref{fig:resourceScreenshot}). The web interface provides a split view of English articles segmented by sentence (right) and corresponding ASL video interpretations (left). To upload a video, contributors select a sentence, which triggers their webcam to start recording and for the camera feed to display on the left-hand side. Contributors can play back recordings and re-record as desired. 
We collected one interpretation per article, with the exception of the five control articles that we asked all interpreters to record. 
Interpreters were assigned to articles according to best fit in terms of skills and expertise.

Once data collection had completed, the research team validated the final set of videos. We automatically removed invalid recordings, manually reviewed a random sample of videos from each contributor, and manually examined length outliers. Removed videos were: 110 with webcam failure, 500 corrupted, 19 with large discrepancies in text and video length (video $<$ 3s, text $\geq5$ words), 1 long outlier (>1000s), and 7 videos shorter than 1s.

\subsection{Dataset annotation}\label{sec:annotation}

To help train fingerspelling detection and alignment models described in \S\ref{sec:task}, and to evaluate their performance, we collected annotations for video segments that contain fingerspelling and the English words being fingerspelled. 
Annotators for this task used our bilingual interface (\autoref{fig:resourceScreenshot}) to view English text and ASL recordings in parallel. 
Annotators were asked to view the requested content, and mark the start and stop times of each occurrence of fingerspelling in a spreadsheet, and label each with the corresponding English word. They were asked to include only strict fingerspelling (e.g. not including numbers or loan signs), and to annotate each word in a contiguous fingerspelled sequence separately. 
All annotators were professional ASL interpreters, distinct from the set of interpreters we used for data collection, and were compensated at a standard hourly remote interpreter rate. 

To establish inter-annotator agreement, we first recruited a set of 5 annotators to all annotate the same 15 sentences, randomly selected to span 15 different articles recorded by 15 different interpreters. 
On this set, we obtain mean pairwise intersection over union (IOU; see \autoref{sec:eval}) of 0.7 
across frame spans containing fingerspelling, indicating a significant inter-annotator agreement (as a random baseline, we computed IOU scores between annotators after randomly shuffling the frame spans for its sentence, and obtain a mean IOU score of 0.15). 


After establishing suitable inter-annotator agreement, additional interpreters annotated a larger set of contents consisting of 15 complete articles: 
three separate interpretations for each of the five control articles. We selected interpretations at random such that each of the 15 recordings contains different interpreters. 
We divide the annotations between 2 annotators who participated in the smaller collection, such that one annotator was assigned all recordings for 2 of the control articles, and the other to all recordings of the other 3 control articles, to reduce the time required to familiarize themselves with new content. 
In total, we collected fingerspelling annotations for 15 articles and 507 sentences.

\subsection{Dataset statistics}\label{sec:stats}

\begin{figure}[t]
    \centering
    \includegraphics[width=\linewidth]{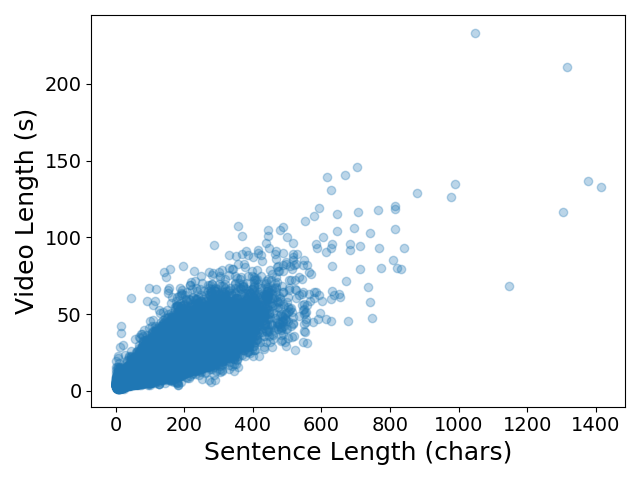}
    \vspace{-2em}
    \caption{Scatterplot of video size. x-axis: sentence length (characters), y-axis: video length (seconds).} 
    \label{fig:videoLengths}
\end{figure}

The final dataset contains 64,266 videos, each corresponding to an English sentence or section title from a STEM-related Wikipedia article. Because the videos were recorded by professional ASL interpreters, they include plain backgrounds, and the interpreters typically wear plain-colored clothes to help with clarity of the signed content. 
The total video content is 315.84 hours, with individual videos ranging from 1.2 to 233 seconds ($\mu= 17.7$, $\sigma=10.3$). 
Participants each contributed between 23 and 10129 videos ($\mu=1737$, $\sigma=2105$). \autoref{fig:videoLengths} provides a plot of sentence length versus video length. 



We estimate the percentage of signs that are fingerspelled in our data by taking the number of English words annotated as fingerspelled, divided by the total number of English words in our annotated subset. Dividing by the total number of English words gives a lower bound on the percentage of fingerspelled signs, since not each word in English is signed in ASL. In addition, we also divide the number of fingerspelled words by the total number of non-stop words. The number of non-stop English words provides a better approximation of the number of ASL signs. 
In \autoref{tab:fs-percent}, we show the percentage of fingerspelling in our data, as well as fingerspelling percentages reported in \citet{morford2003frequency}. We find that in both estimates of fingerspelling in our dataset, there is a high frequency of fingerspelling in our data compared to other domains (casual, formal, narrative, and all signing). This suggests that STEM ASL interpretations have an especially high rate of fingerspelling.

We also manually inspect the types of words that are fingerspelled in a sample of 5 articles (\autoref{tab:fs-type}). We categorize fingerspelled words into 4 categories: technical words related to \textbf{STEM} (e.g. acid, electromagnetism), \textbf{proper nouns} and acronyms (e.g. Fourier, NASA), fingerspelled \textbf{loan words}\footnote{Fingerspelled loan words are ASL signs borrowed from English, where commonly fingerspelled words become more like signs in their own right by undergoing changes that make them faster and easier to produce.} (e.g. \#IF, \#NOV), and \textbf{other} words that do not belong to the previous 3 categories (e.g. `panda', `royalties'). We find that STEM words account for the majority of fingerspelling in our dataset (63.9\%). This suggests that the high rate of fingerspelling in our dataset is mainly attributed to reasons discussed in \autoref{sec:intro}: interpreters have a tendency to fingerspell STEM concepts.


\begin{table}[t]
    \centering
    \begin{tabular}{lr}
    \toprule
      & \textbf{\% Finger-} \\
     \textbf{Domain} & \textbf{spelling} \\
     \midrule
    Casual & 8.7 \\
    Formal & 4.8 \\
    Narrative & 3.3 \\
    Total & 6.4 \\
    \midrule
    \datasetshort{} (all words) & 18.6 \\
    \datasetshort{} (non-stop words) & 31.5 \\
     \bottomrule
    \end{tabular}
    \caption{Percentage of fingerspelling by domain; top rows are reported by \citet{morford2003frequency}.}
    \label{tab:fs-percent}
\end{table}

\begin{table}[t]
    \centering
    \begin{tabular}{lrr}
    \toprule
    & \multicolumn{2}{c}{\textbf{Fingerspelling}} \\
    \cmidrule(lr){2-3}
     \textbf{Category} & \textbf{\#} & \textbf{\%} \\
     \midrule
    STEM & 417 & 63.9 \\
    Proper noun & 137 & 21.0 \\
    Loan word &69 & 10.6 \\
    Other & 30 & 4.6 \\
     \bottomrule
    \end{tabular}
    \caption{Categories of fingerspelled words.}
    \label{tab:fs-type}
\end{table}

\subsection{Use cases}\label{sec:uses}

Given the novel domain of \dataset{}, our dataset can be used for various new studies and challenges. We propose a series of appropriate use cases for our dataset with applications in human-centered natural language processing (NLP).

\paragraph{Automatic sign suggestion.} Our dataset reflects an increased usage of fingerspelling in interpretations of STEM documents. We suggest developing systems to detect when fingerspelling is used and suggest appropriate ASL signs to use instead. Suggestions would be dependent on the domain and context (e.g. ``protein'' in the context of nutrition, structural biology, or protein engineering may have distinct ASL signs), as well as on the audience (e.g. the sign to use for an elementary school class may be different from the sign to use with a college audience). Fingerspelling may be appropriate in some cases as well, for example when introducing a new sign that is not well-known. 

\paragraph{Translationese / interpretese.} Because our dataset is prompted from an English source sentence, it is prone to having effects of \textit{translationese} \cite{koppel-ordan-2011-translationese}, such as English-influenced word order, segmentation of ASL into English sentence boundaries, signs for English homonyms being used instead of the appropriate sign, and increased fingerspelling. We propose training models to detect and repair translationese, as well as conducting potential linguistic studies around interpretese \cite{shlesinger2009towards} of ASL using our dataset. 

\paragraph{Sign variation.} Five of our articles are interpreted by all 37 ASL interpreters in our study. These articles provide a unique opportunity to study variations in how individuals sign and interpret the same English sentence, especially STEM concepts where ASL signs are not stabilized. 


\paragraph{Sign linking / retrieval.} Related to \textit{sign variation}, our dataset contains examples of English words that may be interpreted differently across interpreters and context. This data can be used to train models that links different versions of ASL signs for the same concept (e.g. one interpreter may sign ``electromagnetism'' using the signs for ELECTRICITY and MAGNET, another interpreter may interpret the same word using a sign that visually describes an electromagnetic field).

\paragraph{Automatic STEM translation.} Our dataset can be used to train, fine-tune, and/or evaluate model capabilities in translating technical content from English to ASL. Technically, our dataset could be used to develop models to translate from ASL to English, however, this direction is not preferred since our dataset contains interpreted ASL which may differ from unprompted ASL \cite{shlesinger2009towards}.





\section{Automatic sign suggestion}\label{sec:task}

As an initial step towards exploring our dataset for modeling and downstream applications, we focus on \textbf{automatic sign suggestion}. 
This task can be broken into two separate subtasks: identifying fingerspelled words, and retrieving videos of those words being signed in a dictionary. We focus on modeling the first aspect---given an English sentence and its ASL interpretation, we detect all instances of fingerspelling (\textbf{fingerspelling detection}) and align each detected fingerspelling segment to the word in the English sentence that it spells out (\textbf{fingerspelling alignment})---and leave the retrieval task to future work. 


\subsection{Formal task specification}

\paragraph{Fingerspelling detection.}
Given an input ASL video $x$, represented as a sequence of frames $\{ x_1, x_2, \dots, x_n \}$, the goal is to output a set of \textit{frame spans} $\mathcal{F}$ that correspond to instances of finger spelling: $\mathcal{F} = \{ [s_1, e_1], [s_2, e_2], \dots, [s_k, e_k] \}$, where frame ranges $x_{s_i}$ through $x_{e_i}$ correspond to fingerspelling for all $i$.



\paragraph{Fingerspelling alignment.}
Given an input ASL video $x$, a corresponding English sentence $w = \{ w_1, w_2, \dots, w_m \}$, and a set of frame spans $\mathcal{F}$, the goal is to associate with each span $[s_i, e_i] \in \mathcal{F}$ with an index $j_i \in \{ 1, \dots, m \}$ such that the word fingerspelled in $[x_{s_i}, x_{s_i}]$ corresponds to the English word $w_{j_i}$ in the sentence.





\section{Models for fingerspelling detection and alignment}\label{sec:model}

We now propose baseline systems for fingerspelling detection and alignment: a neural model on \dataset{} for fingerspelling detection and a heuristic model based on word frequency for fingerspelling alignment. 

\subsection{Preprocessing}\label{sec:preprocessing}

We use 2D human pose representations extracted from raw ASL video using MediaPipe \cite{zhang2020mediapipe} with model complexity 1 and minimum detection confidence 0.5. We select 75 keypoints of the two hands and body, and discard face mesh keypoints to reduce dimensionality. We also discard the z-coordinate of each keypoint because depth estimation in MediaPipe may not be accurate. 

\begin{figure}[t]
    \centering
    \includegraphics[width=0.7\linewidth]{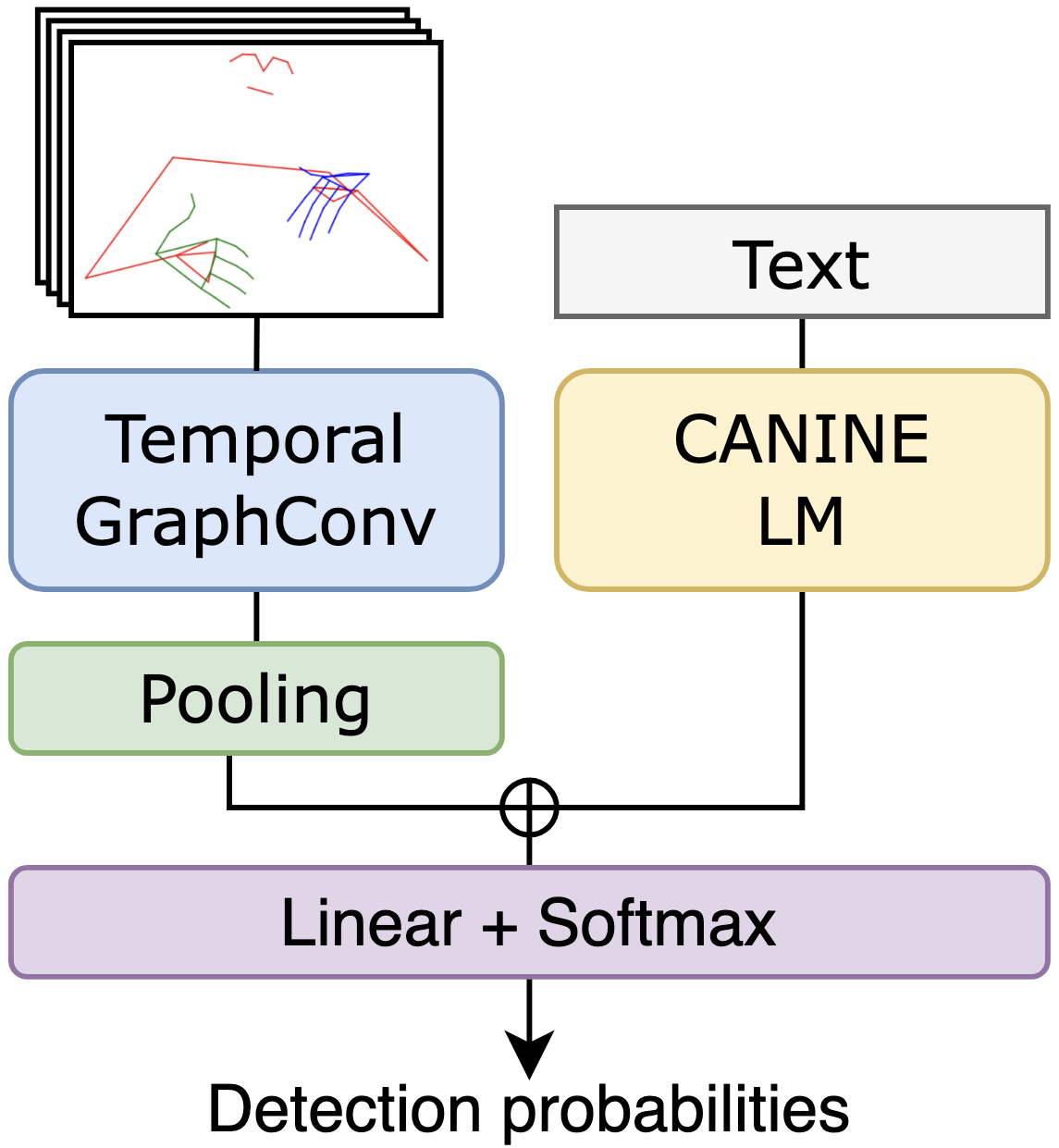}
    \caption{Fingerspelling detection model. Frames of ASL keypoints are processed by a temporal graph convolutional network, and the associated English sentence is processed by the CANINE pre-trained language model. The two representations are concatenated then passed to a linear layer to predict fingerspelling frames.}
    \label{fig:model}
\end{figure}

\subsection{Fingerspelling detection}

First, we pre-train models with contrastive learning to learn representations of English-ASL data on unlabeled data. Then, we train models on fingerspelling annotations collected in \S\ref{sec:annotation} to perform fingerspelling detection. 


\subsubsection{Model architecture}
Our model consists of a video and a text component (\autoref{fig:model}).
We use a Pytorch \cite{paszke2017automatic} implementation of the temporal graph convolutional network from \citet{li2020word} for the video component of the model. Our network stacks 6 graph convolution blocks with hidden size 64. Each input video has dimension 75, and are padded or truncated to 2000 frames (just over one minute). 

For the text component, we use CANINE \cite{clark2022canine}, a pre-trained character-level Transformer \cite{vaswani2017attention} language model, based on the intuition that character-level processing may be better suited for fingerspelling. We use the HuggingFace \cite{wolf2019huggingface} implementation (\texttt{google/canine-c}). CANINE has 12 layers, 12 attention heads, and hidden size 768. We pad or truncate each sentence to 600 characters.

The ASL video is fed to the video component, the associated English sentence is fed to the text component. The output representations of the two components are concatenated, then passed to a linear layer that outputs probabilities of frames containing fingerspelling.

\subsubsection{Self-supervised pretraining}

Our model is first pre-trained in a self-supervised manner for representation learning. This leverages the English-ASL data in \dataset{} without requiring explicit labels for fingerspelling. We define two contrastive learning objectives aimed to learn representations of ASL videos, and associations between ASL video and English text:

\begin{itemize}
\item \textbf{Temporal contrastive learning objective \cite{hu2021contrast}:}
Given two randomly sampled ASL video clips $x_1, x_2$ each of length 200 frames, the model predicts whether the two clips come from the same video, and if so, whether $x_1$ happens earlier or later than $x_2$ in the video. In examples where $x_1$ and $x_2$ are sampled from the same video, we ensure that the original video is longer than 400 frames and that there is no overlap between the two clips. We balance the number of samples in each class and we optimize for cross-entropy loss on the trinary prediction.

\item \textbf{Sentential contrastive learning objective:}
We define a new contrastive learning objective to learn representations between ASL videos and text. Given an ASL video $x$ of length 2000 frames and two English sentences $w^1, w^2$ each of length 300 characters, the model predicts which English sentence the ASL video $x$ is an interpretation of. Since we take a shorter sentence length, some sentences are truncated, so the model also learns to map whether a subset of frames in $x$ matches a sentence. We balance the number of samples in each class and we optimize for binary cross-entropy loss.
\end{itemize}

\subsubsection{Supervised training}
 On top of the initialized weights from pretraining, we fine-tune our model for fingerspelling detection using the subset of the dataset annotated with fingerspelling labels. We represent labels as a sequence $y = y_1,y_2,...,y_n$ where $n$ is the number of frames in the input video, $y_i = 1$ if the $i$-th frame is part of a fingerspelling segment, and $y_i = 0$ otherwise. We use weighted binary cross-entropy loss to balance the 0 and 1 labels.

 We pretrain for 50 epochs and fine-tune for 20 epochs. To measure the contribution of pretraining, we also train a model from random initialization for 40 epochs. For both pretraining and supervised training, we use Adam optimizer \cite{kingma2014adam} (lr=0.001) and batch size 32. Each model was trained on one RTX A6000 GPU for 45 hours.

\subsection{Fingerspelling alignment}
For fingerspelling alignment, we use a heuristic approach based on the property that fingerspelling is often used to spell infrequent English words \cite{battison1978lexical}. We first estimate the frequency of English words using 10,000 randomly sampled Wikipedia articles in English from all domains (not necessarily STEM). Then, given a video where $n$ fingerspelling segments are detected, we predict the $n$ least frequent words in the English sentence as the words fingerspelled in the video segments. The first fingerspelling segment is aligned to the first predicted English word, and so on.

\subsection{Evaluation}\label{sec:eval}
Given the limited annotated data for fine-tuning and evaluation, we use cross-validation to evaluate models. The subset with fingerspelling annotations contains 5 articles, we use 4 articles to fine-tune and 1 article to evaluate the model, and repeat this process 5 times changing the article used for evaluation each time. We report the average performance over samples in all 5 articles. We also evaluate detection and alignment independently to better understand how models perform on each step.

To measure detection and alignment accuracy, we compute the intersection over union (IOU). Given a ground truth set of frame spans $\mathcal{F}$ and corresponding alignment indices $j$, and a predicted set of frame spans $\mathcal{F}'$ and alignment indices $j'$, we compute $IOU(F,F') = | F \cap F' | ~/~ | F \cup F' |$, where 
$F = \{ (f_k, w_{j_i}) : [s_i,e_i] \in \mathcal{F}, k \in [s_i,e_i] \}$
is the set of frames (and paired English words) from $\mathcal{F}$, and $F'$ is computed equivalently for $\mathcal{F}'$. For evaluating detection alone, we use the same score but ignore the aligned English words.



\begin{table}[t]
    \centering
    \begin{tabular}{lccc}
    \toprule
    && \multicolumn{2}{c}{\textbf{Our Model}} \\
    \cmidrule(lr){3-4}
      & \textbf{Random}  & \textbf{w/o PT} & \textbf{w/ PT} \\
     \midrule
    \textbf{Detection} & 0.06 & 0.19 & 0.28 \\
    \textbf{Alignment} & 0.06 & \multicolumn{2}{c}{0.13} \\
     \bottomrule
    \end{tabular}
    \caption{Mean IOU scores on fingerspelling detection and alignment. Our models outperform the random baseline, and pre-training (PT) with contrastive learning improves fingerspelling detection performance.}
    \label{tab:results}
\end{table}

\section{Results \& discussion}\label{sec:discussion}
\subsection{Results}
We report IOU scores for fingerspelling detection and alignment in \autoref{tab:results}. As a random baseline for fingerspelling detection, we compute the percentage of frames $P$ containing fingerspelling on a held-out dataset. Then, for each frame in the input video, we predict that it is part of a fingerspelling span with probability $P$. Similarly for fingerspelling alignment, we estimate the percentage of letters $P'$ that is fingerspelled in the video, and predict each letter in the input sample that it is fingerspelled with probability $P'$.

We find that both baseline models with and without pre-training have a higher IOU than random. In addition, pre-training with contrastive learning improves model performance on detection. Nonetheless, our baselines obtain low IOU scores on both tasks, which underscores the potential of this task as a good, challenging benchmark for future models in sign language understanding. 

\subsection{Error analysis \& discussion}
We conduct an error analysis of our fingerspelling detection and alignment models by manually inspecting failure cases to better understand the performance limitations and potential areas for improvement.

For fingerspelling detection, our model frequently exhibits false positives on instances of one-handed ASL signs with rapid hand movements. On the other hand, the model struggles with false negatives in instances involving very short fingerspelling words (e.g. two-letter abbreviations), or in dynamic contexts where there is a seamless transition between ASL signing and fingerspelling, especially in fast-paced signing, so the model struggles to isolate the fingerspelling segments. 

For fingerspelling alignment, the main source of error comes from a difference in the order in which different fingerspelled words appear in the ASL video and the English sentence. Our heuristic method makes the assumption that the fingerspelled words in ASL and English appear in the same order, which is not always true due to differences in syntax between the two languages. Therefore, a heuristic that takes into account the syntax of English and ASL, or a model trained to perform fingerspelling alignment, could improve performance. Another common error is when a fingerspelling clip contains a compound word or multiple words, but only one word from the English sentence is mapped to the clip.

We also believe that lack of data with fingerspelling annotations is a bottleneck of this challenge, for both fingerspelling detection and alignment. While our presented dataset provides a large amount of video data, it does not include annotations over all videos. To improve model performance, future work can expand dataset size by either collecting and annotating more data, performing data augmentation, or using synthetic data, especially targeting the type of data where our model has often made errors as described above. We also find evidence of contrastive learning improving fingerspelling detection, therefore we also suggest exploring other unsupervised and semi-supervised objectives for pre-training.




\section{Conclusion}

We present \dataset{}: a large dataset of Wikipedia articles on STEM topics professionally interpreted from English to ASL. The first continuous sign language STEM dataset, \dataset{} captures unique characteristics of interpreted STEM content and serves as a valuable resource for developing AI tools that enhance the accessibility of STEM education for deaf students. We also present baseline models working towards \challenge, which highlight the challenge of this new task while also suggesting that contrastive learning may be a promising approach. We invite others to use \dataset{} to further advance automatic sign suggestion and address other new challenges in human-centered NLP.

\section*{Limitations}

We discussed how interpreters may resort to suboptimal strategies when interpreting STEM documents, such as frequent fingerspelling in \autoref{sec:bg}. As a result, videos in \dataset{} may not be suitable for training a generative ASL model since it is not un-prompted natural signing and may have noticeable translationese effects from English. We, therefore, discuss appropriate use cases of our dataset with these properties in mind in \autoref{sec:uses}, and also leverage these features to study how AI systems can assist humans on difficult interpretations to overcome these limitations.

For dataset collection, we recruited interpreters from an accredited interpreting services provider. We initially recruited certified deaf interpreters (CDI), however, the volume of interpretations became too large for the CDI pool. We therefore also included certified hearing ASL interpreters. Although all interpreters are RID and DBE certified, the skill and fluency of signing may vary between interpreters included in our dataset.

We provide baseline models of fingerspelling detection and alignment that can be later used to query for ASL signs from a lexicon, with the motivation to use this pipeline for automatic sign suggestion. However, as discussed in \S\ref{sec:uses}, a full, working system for automatic sign suggestion would require fine-grained understanding of the domain and context to suggest signs that are appropriate to the document and the audience, which exceeds current resources. The modeling framework we use is useful as a simpler initial step towards exploring automatic sign suggestion, but is not the only or the long-term approach for addressing this task.

\section*{Acknowledgements}

We thank all dataset participants for contributing, and the interpreter managers for facilitating the collection. We also thank Mary Bellard, William Thies, Philip Rosenfield, Sharon Gillett, Paul Oka, and Abraham Glasser for meaningful discussions, formative work, and support.

\bibliography{anthology,custom}
\bibliographystyle{acl_natbib}

\newpage
\appendix

\section{Datasheet}
\label{sec:datasheet}

We upload the datasheet for the \dataset{} dataset in the following pages.





\begin{center}
    \Huge
    Datasheet for \\ ASL STEM Wiki  \\
\end{center}


\definecolor{darkblue}{RGB}{46,25, 110}

\newcommand{\dssectionheader}[1]{%
   \noindent\framebox[\columnwidth]{%
      {\fontfamily{phv}\selectfont \textbf{\textcolor{darkblue}{#1}}}
   }
}

\newcommand{\dsquestion}[1]{%
    {\noindent \fontfamily{phv}\selectfont \textcolor{darkblue}{\textbf{#1}}}
}

\newcommand{\dsquestionex}[2]{%
    {\noindent \fontfamily{phv}\selectfont \textcolor{darkblue}{\textbf{#1} #2}}
}

\newcommand{\dsanswer}[1]{%
   {\noindent #1 \medskip}
}

\dssectionheader{Motivation}

\dsquestionex{For what purpose was the dataset created?}{Was there a specific task in mind? Was there a specific gap that needed to be filled? Please provide a description.}

\dsanswer{This dataset was created for two purposes: 1) to enable a small bilingual informational resource in both English and ASL, and 2) to provide continuous labelled sign language data for research. There is a severe shortage of such publicly available data, which is a primary barrier to research and technology advancement. More information about the bilingual resource can be found in a prior publication \cite{glasser2022asl}, and on the prototype website 
\url{https://aslgames.azurewebsites.net/wiki/}.

Additional background and all supplementary materials, including the bilingual website link, are available on the project page website \url{https://www.microsoft.com/en-us/research/project/asl-stem-wiki/}. 
}

\dsquestion{Who created this dataset (e.g., which team, research group) and on behalf of which entity (e.g., company, institution, organization)?}

\dsanswer{This dataset was created by Microsoft Research.
}

\dsquestionex{Who funded the creation of the dataset?}{If there is an associated grant, please provide the name of the grantor and the grant name and number.}

\dsanswer{Microsoft funded the creation of the dataset.
}

\dsquestion{Any other comments?}

\dsanswer{N/A
}

\bigskip
\dssectionheader{Composition}

\dsquestionex{What do the instances that comprise the dataset represent (e.g., documents, photos, people, countries)?}{ Are there multiple types of instances (e.g., movies, users, and ratings; people and interactions between them; nodes and edges)? Please provide a description.}

\dsanswer{The dataset consists of continuous American Sign Language (ASL) videos, with associated metadata. The contents are interpretations of English Wikipedia articles related to STEM topics, recorded by professional ASL interpreters. Each recording corresponds to a single English sentence or section title in the original text. The videos are provided with the corresponding segmented English texts. 
}

\dsquestion{How many instances are there in total (of each type, if appropriate)?}

\dsanswer{The dataset consists of 64,266 videos, each corresponding to an English sentence or section title from a STEM-related Wikipedia article. 
}

\dsquestionex{Does the dataset contain all possible instances or is it a sample (not necessarily random) of instances from a larger set?}{ If the dataset is a sample, then what is the larger set? Is the sample representative of the larger set (e.g., geographic coverage)? If so, please describe how this representativeness was validated/verified. If it is not representative of the larger set, please describe why not (e.g., to cover a more diverse range of instances, because instances were withheld or unavailable).}

\dsanswer{The dataset is a sample. It contains a sample of English-to-ASL sentence-by-sentence translations of STEM-related articles from Wikipedia, and the videos contain a sample of 37 professional ASL interpreters.
}

\dsquestionex{What data does each instance consist of? “Raw” data (e.g., unprocessed text or images) or features?}{In either case, please provide a description.}

\dsanswer{Each data point consists of a video. Each video contains a single professional ASL interpreter executing an ASL translation of an English sentence or section heading from a STEM-related Wikipedia article.
}

\dsquestionex{Is there a label or target associated with each instance?}{If so, please provide a description.}

\dsanswer{Yes, each video corresponds to a portion of a Wikipedia article. We provide the corresponding text.
}

\dsquestionex{Is any information missing from individual instances?}{If so, please provide a description, explaining why this information is missing (e.g., because it was unavailable). This does not include intentionally removed information, but might include, e.g., redacted text.}

\dsanswer{No.
}

\dsquestionex{Are relationships between individual instances made explicit (e.g., users’ movie ratings, social network links)?}{If so, please describe how these relationships are made explicit.}

\dsanswer{Yes, videos are related to one another, in that they correspond to sequential sentences or titles in Wikipedia articles. We provide this sequential information in the text metadata.
}

\dsquestionex{Are there recommended data splits (e.g., training, development/validation, testing)?}{If so, please provide a description of these splits, explaining the rationale behind them.}

\dsanswer{In our paper accompanying the dataset release, we used the following splits: 
\begin{itemize}
    \item training: all recordings of the five control articles (Acid catalysis, EDGE species, Hal Anger (person), Relativistic electromagnetism, Standard score), which provided multiple recordings per sentence
    \item test: all remaining articles, which provided a single recording per sentence
\end{itemize}

Dataset users can consider splitting the dataset by article or interpreter, to help preserve the independence of the test set.

}

\dsquestionex{Are there any errors, sources of noise, or redundancies in the dataset?}{If so, please provide a description.}

\dsanswer{Though the translations were made by professional ASL interpreters, the translations are still human-generated, and may contain errors. Because the translations were made from English to ASL, the former language likely influences the secondary language (e.g. in grammatical structures). The English text was also segmented into sentence units, forcing the translation to occur sentence-by-sentence, which further constrained the flexibility and naturalness of the ASL. Additionally, because the content is STEM-related and sometimes technical, fingerspelling is often used to represent concepts where signs do not exist or may have been unfamiliar to the interpreter.

The dataset is also missing occasional sentences from the original Wikipedia articles. While the ASL interpreters were asked to record translations of entire Wikipedia articles, occasionally portions were skipped, and some additional videos were removed during cleaning. 
To validate the data, the research team manually reviewed a random sample of videos from each contributor, ran scripts to check for invalid recordings, and manually examined outliers. Removed videos were: 110 with webcam failure, 500 corrupted, 19 with large discrepancies in text and recording length (video < 3s, text $\geq5$ words), 1 large outlier (>1000s), and 7 shorter than 1s.
}

\dsquestionex{Is the dataset self-contained, or does it link to or otherwise rely on external resources (e.g., websites, tweets, other datasets)?}{If it links to or relies on external resources, a) are there guarantees that they will exist, and remain constant, over time; b) are there official archival versions of the complete dataset (i.e., including the external resources as they existed at the time the dataset was created); c) are there any restrictions (e.g., licenses, fees) associated with any of the external resources that might apply to a future user? Please provide descriptions of all external resources and any restrictions associated with them, as well as links or other access points, as appropriate.}

\dsanswer{The videos correspond to English Wikipedia articles. We provide the mapping between ASL videos and English text. The Wikipedia text has been published under a Creative Contents license, and is available for public download (\url{https://en.wikipedia.org/wiki/Wikipedia:Database_download}, July 2020).
}

\dsquestionex{Does the dataset contain data that might be considered confidential (e.g., data that is protected by legal privilege or by doctor-patient confidentiality, data that includes the content of individuals non-public communications)?}{If so, please provide a description.}

\dsanswer{No, the data is not confidential. All participants consented to providing recordings and public release of the dataset.
}

\dsquestionex{Does the dataset contain data that, if viewed directly, might be offensive, insulting, threatening, or might otherwise cause anxiety?}{If so, please describe why.}

\dsanswer{No, we do not expect contents to be offensive. The topic is STEM.
}

\dsquestionex{Does the dataset relate to people?}{If not, you may skip the remaining questions in this section.}

\dsanswer{Yes, the videos are of sign language interpreters.
}

\dsquestionex{Does the dataset identify any subpopulations (e.g., by age, gender)?}{If so, please describe how these subpopulations are identified and provide a description of their respective distributions within the dataset.}

\dsanswer{The people in the videos are professional ASL interpreters.
}

\dsquestionex{Is it possible to identify individuals (i.e., one or more natural persons), either directly or indirectly (i.e., in combination with other data) from the dataset?}{If so, please describe how.}

\dsanswer{Yes, it is possible to identify individuals in the dataset, since the interpreters' faces and upper bodies are captured in the videos.
}

\dsquestionex{Does the dataset contain data that might be considered sensitive in any way (e.g., data that reveals racial or ethnic origins, sexual orientations, religious beliefs, political opinions or union memberships, or locations; financial or health data; biometric or genetic data; forms of government identification, such as social security numbers; criminal history)?}{If so, please provide a description.}

\dsanswer{The data may be considered sensitive, in that the interpreters' faces and upper bodies are captured in the videos.
}

\dsquestion{Any other comments?}

\dsanswer{N/A
}

\bigskip
\dssectionheader{Collection Process}

\dsquestionex{How was the data associated with each instance acquired?}{Was the data directly observable (e.g., raw text, movie ratings), reported by subjects (e.g., survey responses), or indirectly inferred/derived from other data (e.g., part-of-speech tags, model-based guesses for age or language)? If data was reported by subjects or indirectly inferred/derived from other data, was the data validated/verified? If so, please describe how.}

\dsanswer{Professional interpreters recorded each video, which was automatically linked to the corresponding English text. This process was facilitated by a web platform that the research team created for this collection (described subsequently).
}

\dsquestionex{What mechanisms or procedures were used to collect the data (e.g., hardware apparatus or sensor, manual human curation, software program, software API)?}{How were these mechanisms or procedures validated?}

\dsanswer{The data was collected through a website created explicitly for this dataset collection. The platform provides a full English text on the right side of the screen. The interpreters have full access to the text, and can progress through the text and record themselves providing a translation in the website. A record button is available, which triggers recording through their computer's webcam. The interface also supported playing back recordings and re-recording. The resulting translations are available for content consumers to view, for example to enable access to spot-translations to improve article accessibility to people whose primary language is ASL. The full platform design is described in \cite{glasser2022asl}.
}

\dsquestion{If the dataset is a sample from a larger set, what was the sampling strategy (e.g., deterministic, probabilistic with specific sampling probabilities)?}

\dsanswer{N/A
}

\dsquestion{Who was involved in the data collection process (e.g., students, crowdworkers, contractors) and how were they compensated (e.g., how much were crowdworkers paid)?}

\dsanswer{Microsoft Research designed and built the collection platform. Professional interpreters provided recorded translations through the web platform. The interpreters were paid at standard hourly ASL interpreter rates for virtual interpretation jobs.
}

\dsquestionex{Over what timeframe was the data collected? Does this timeframe match the creation timeframe of the data associated with the instances (e.g., recent crawl of old news articles)?}{If not, please describe the timeframe in which the data associated with the instances was created.}

\dsanswer{The videos were collected between March and June 2022.
}

\dsquestionex{Were any ethical review processes conducted (e.g., by an institutional review board)?}{If so, please provide a description of these review processes, including the outcomes, as well as a link or other access point to any supporting documentation.}

\dsanswer{Yes, the project was reviewed by Microsoft's Institutional Review Board (IRB). The collection platform and the dataset release also underwent additional Ethics and Compliance reviews by Microsoft. 
}

\dsquestionex{Does the dataset relate to people?}{If not, you may skip the remaining questions in this section.}

\dsanswer{Yes, the dataset consists of videos of human ASL interpreters.
}

\dsquestion{Did you collect the data from the individuals in question directly, or obtain it via third parties or other sources (e.g., websites)?}

\dsanswer{The videos were recorded by the ASL interpreters.
}

\dsquestionex{Were the individuals in question notified about the data collection?}{If so, please describe (or show with screenshots or other information) how notice was provided, and provide a link or other access point to, or otherwise reproduce, the exact language of the notification itself.}

\dsanswer{Yes, the interpreters controlled the recording process (e.g. start, stop, re-recording). A screenshot of the recording interface is provided in Figure 3 of \cite{glasser2022asl}. Contributors also engaged in a consent process prior to contributing any data. 
}

\dsquestionex{Did the individuals in question consent to the collection and use of their data?}{If so, please describe (or show with screenshots or other information) how consent was requested and provided, and provide a link or other access point to, or otherwise reproduce, the exact language to which the individuals consented.}

\dsanswer{Yes, participants engaged in a consent process through the web platform prior to contributing.  For the exact consent text, please visit \url{https://www.microsoft.com/en-us/research/project/asl-stem-wiki/}.
}

\dsquestionex{If consent was obtained, were the consenting individuals provided with a mechanism to revoke their consent in the future or for certain uses?}{If so, please provide a description, as well as a link or other access point to the mechanism (if appropriate).}

\dsanswer{Yes, participants could contact the research team directly, and could delete any of their recordings through the web platform.
}

\dsquestionex{Has an analysis of the potential impact of the dataset and its use on data subjects (e.g., a data protection impact analysis) been conducted?}{If so, please provide a description of this analysis, including the outcomes, as well as a link or other access point to any supporting documentation.}

\dsanswer{Yes, a Data Protection Impact Analysis (DPIA) has been conducted, including taking a detailed inventory of the data types collected and stored and retention policy, and was successfully reviewed by Microsoft.
}

\dsquestion{Any other comments?}

\dsanswer{N/A
}

\bigskip
\dssectionheader{Preprocessing/cleaning/labeling}

\dsquestionex{Was any preprocessing/cleaning/labeling of the data done (e.g., discretization or bucketing, tokenization, part-of-speech tagging, SIFT feature extraction, removal of instances, processing of missing values)?}{If so, please provide a description. If not, you may skip the remainder of the questions in this section.}

\dsanswer{To validate the data, the research team manually reviewed a random sample of videos from each contributor, ran scripts to check for invalid recordings, and manually examined outliers. Removed videos were: 110 with webcam failure, 500 corrupted, 19 with large discrepancies in text and recording length (video < 3s, text $\geq5$ words),  1 large outlier (>1000s), and 7 shorter than 1s.
}

\dsquestionex{Was the “raw” data saved in addition to the preprocessed/cleaned/labeled data (e.g., to support unanticipated future uses)?}{If so, please provide a link or other access point to the “raw” data.}

\dsanswer{No, not publicly. 
}

\dsquestionex{Is the software used to preprocess/clean/label the instances available?}{If so, please provide a link or other access point.}

\dsanswer{No, but these procedures are easily reproducible using public software. 
}

\dsquestion{Any other comments?}

\dsanswer{N/A
}

\bigskip
\dssectionheader{Uses}

\dsquestionex{Has the dataset been used for any tasks already?}{If so, please provide a description.}

\dsanswer{Yes, we provide fingerspelling detection and alignment baselines in the accompanying paper publication.
}

\dsquestionex{Is there a repository that links to any or all papers or systems that use the dataset?}{If so, please provide a link or other access point.}

\dsanswer{Yes, the link is available on our project page at  \url{https://www.microsoft.com/en-us/research/project/asl-stem-wiki/}.
}

\dsquestion{What (other) tasks could the dataset be used for?}

\dsanswer{The dataset can be used for a range of tasks, including:

Fingerspelling detection and recognition - Our dataset reflects an increased usage of fingerspelling in interpretations of STEM documents. Identifying instances of fingerspelling and mapping these instances onto the corresponding English words can help researchers better understand signing patterns. We provide benchmarks for fingerspelling detection and recognition in our paper accompanying the dataset release. The ability to detect and recognize fingerspelling can also enable richer downstream applications, such as automatic sign suggestion (described subsequently).

Automatic sign suggestion - Also motivated by the prevalence of fingerspelling in STEM interpretations, we suggest developing systems to detect when fingerspelling is used and suggest appropriate ASL signs to use instead. Suggestions would be dependent on the domain and context (e.g. ``protein'' in the context of nutrition, structural biology, or protein engineering may have distinct ASL signs), as well as on the audience (e.g. the sign to use for an elementary school class may be different from the sign to use with a college audience). Fingerspelling may be appropriate in some cases as well, for example when introducing a new sign that is not well-known.

Translationese/Interpretese - Because our dataset is prompted from an English source sentence, it is prone to having effects of translationese \cite{koppel2011translationese}, such as English-influenced word order, segmentation of ASL into English sentence boundaries, signs for English homonyms being used instead of the appropriate sign, and increased fingerspelling. We propose training models to detect and repair translationese, as well as potential translation and interpretation studies around interpretese \cite{shlesinger2009towards} of ASL.

Sign variation - Five of our articles are interpreted by all 37 ASL interpreters in our study. These articles provide a unique opportunity to study variations in how individuals sign and interpret the same English sentence, especially STEM concepts where ASL signs are not stabilized.

Sign linking/retrieval - Related to sign variation, our dataset contains examples of English words that may be interpreted differently across interpreters and context. This data can be used to train models that links different versions of ASL signs for the same concept (e.g. one interpreter may sign ``electromagnetism'' using the signs for ELECTRICITY and MAGNET, another interpreter may interpret the same word using a sign that visually describes an electromagnetic field).

Automatic STEM translation - Our dataset can be used to train, fine-tune, and/or evaluate model capabilities in translating technical content from English to ASL. Technically, our dataset could be used to develop models to translate from ASL to English, however, this direction is not preferred since our dataset contains interpreted ASL which may differ from unprompted ASL \cite{shlesinger2009towards}.

}

\dsquestionex{Is there anything about the composition of the dataset or the way it was collected and preprocessed/cleaned/labeled that might impact future uses?}{For example, is there anything that a future user might need to know to avoid uses that could result in unfair treatment of individuals or groups (e.g., stereotyping, quality of service issues) or other undesirable harms (e.g., financial harms, legal risks) If so, please provide a description. Is there anything a future user could do to mitigate these undesirable harms?}

\dsanswer{Dataset users should be aware that the dataset does not consist of natural ASL content. Because the translations were made from English to ASL, the former language likely influences the secondary language (e.g. in grammatical structures). The English text was also segmented into sentence units, forcing the translation to occur sentence-by-sentence, which further constrained the flexibility and naturalness of the ASL. Additionally, because the content is STEM-related and sometimes technical, fingerspelling is often used to represent concepts where signs do not exist or may have been unfamiliar to the interpreter.

It is possible that some of these limitations may be reduced in the future by post-processing or correcting the videos, by combining this dataset with other datasets that do consist of natural ASL-first contents, or by incorporating linguistic knowledge in modeling. Involving fluent ASL team members in key roles in projects can help mediate risks, for example to provide feedback on signing quality, community needs and perspectives, and other relevant guidance and information.
}

\dsquestionex{Are there tasks for which the dataset should not be used?}{If so, please provide a description.}

\dsanswer{As described above, this dataset is not an example of natural ASL-first signing. As a result, we do not recommend using this dataset alone to understand or model natural ASL-first signing. At a minimum, this dataset would need to be used in conjunction with other datasets and/or domain knowledge about sign language in order to more accurately model naturalistic ASL.

More generally, we recommend using this data with meaningful involvement from Deaf community members in leadership roles with decision-making authority at every step from conception to execution. As described in other works, research and development of sign language technologies that involves Deaf community members increases the quality of the work, and can help to ensure technologies are relevant and wanted. Historically, projects developed without meaningful Deaf involvement have not been well received \cite{gloves} and have damaged relationships between technologists and deaf communities.  

We ask that this dataset is used with an aim of making the world more equitable and just for deaf people, and with a commitment to ``do no harm''. In that spirit, this dataset should not be used to develop technology that purports to replace sign language interpreters, fluent signing educators, and/or other hard-won accommodations for deaf people.

}

\dsquestion{Any other comments?}

\dsanswer{N/A
}

\bigskip
\dssectionheader{Distribution}

\dsquestionex{Will the dataset be distributed to third parties outside of the entity (e.g., company, institution, organization) on behalf of which the dataset was created?}{If so, please provide a description.}

\dsanswer{Yes, the dataset is released publicly, to help advance research and development related to sign language.
}

\dsquestionex{How will the dataset will be distributed (e.g., tarball on website, API, GitHub)}{Does the dataset have a digital object identifier (DOI)?}

\dsanswer{The dataset is publicly available for download through the Microsoft Download Center. Links are available through the project page at \url{https://www.microsoft.com/en-us/research/project/asl-stem-wiki/}.

}

\dsquestion{When will the dataset be distributed?}

\dsanswer{The dataset was released on 10/15/2024.
}

\dsquestionex{Will the dataset be distributed under a copyright or other intellectual property (IP) license, and/or under applicable terms of use (ToU)?}{If so, please describe this license and/or ToU, and provide a link or other access point to, or otherwise reproduce, any relevant licensing terms or ToU, as well as any fees associated with these restrictions.}

\dsanswer{Yes, the dataset will be published under a license that permits use for research purposes. The license is provided at \url{https://www.microsoft.com/en-us/research/project/asl-stem-wiki/dataset-license/}.
}

\dsquestionex{Have any third parties imposed IP-based or other restrictions on the data associated with the instances?}{If so, please describe these restrictions, and provide a link or other access point to, or otherwise reproduce, any relevant licensing terms, as well as any fees associated with these restrictions.}

\dsanswer{The Wikipedia article texts are multi-licensed under the Creative Commons Attribution-ShareAlike 3.0 License (CC-BY-SA) and the GNU Free Documentation License (GFDL).
}

\dsquestionex{Do any export controls or other regulatory restrictions apply to the dataset or to individual instances?}{If so, please describe these restrictions, and provide a link or other access point to, or otherwise reproduce, any supporting documentation.}

\dsanswer{No.
}

\dsquestion{Any other comments?}

\dsanswer{N/A
}

\bigskip
\dssectionheader{Maintenance}

\dsquestion{Who will be supporting / hosting / maintaining the dataset?}

\dsanswer{The dataset will be hosted on Microsoft Download Center.
}

\dsquestion{How can the owner / curator / manager of the dataset be contacted (e.g., email address)?}

\dsanswer{Please contact \texttt{ASL\_Citizen@microsoft.com} with any questions.
}

\dsquestionex{Is there an erratum?}{If so, please provide a link or other access point.}

\dsanswer{A public-facing website is associated with the dataset (see \url{https://www.microsoft.com/en-us/research/project/asl-stem-wiki/}). 
We will link to erratum on this website if necessary. 
}

\dsquestionex{Will the dataset be updated (e.g., to correct labeling errors, add new instances, delete instances)?}{If so, please describe how often, by whom, and how updates will be communicated to users (e.g., mailing list, GitHub)?}

\dsanswer{If updates are necessary, we will update the dataset. We will release our dataset with a version number, to distinguish it with any future updated versions.
}

\dsquestionex{If the dataset relates to people, are there applicable limits on the retention of the data associated with the instances (e.g., were individuals in question told that their data would be retained for a fixed period of time and then deleted)?}{If so, please describe these limits and explain how they will be enforced.}

\dsanswer{The dataset will be left up indefinitely, to maximize utility to research. Participants were informed that their contributions might be released in a public dataset.
}

\dsquestionex{Will older versions of the dataset continue to be supported/hosted/maintained?}{If so, please describe how. If not, please describe how its obsolescence will be communicated to users.}

\dsanswer{All versions of the dataset will be released with a version number on Microsoft Download Center to enable differentiation.
}

\dsquestionex{If others want to extend/augment/ build on/contribute to the dataset, is there a mechanism for them to do so?}{If so, please provide a description. Will these contributions be validated/verified? If so, please describe how. If not, why not? Is there a process for communicating/distributing these contributions to other users? If so, please provide a description.}

\dsanswer{We do not have a mechanism for others to contribute to our dataset directly. However, others could create comparable datasets by recording ASL translations of English texts.
}

\dsquestion{Any other comments?}

\dsanswer{N/A
}


\end{document}